\documentclass[journal]{IEEEtran}
\ifCLASSINFOpdf
\else
\fi

\usepackage{graphicx}
\usepackage{caption}
\usepackage{subcaption}
\usepackage{algorithm}
\usepackage{algorithmicx}
\usepackage{amsfonts}
\usepackage{makecell}
\usepackage{amsmath}
\usepackage{algpseudocode}
\usepackage{multicol}
\usepackage{tikz}
\usepackage{pgfplots}
\usepackage{array}
\usepackage{amssymb} % Allows the use of \toprule, \midrule and \bottomrule in tables
\usepackage{float}
\usepackage{mathtools, cuted}
\usepackage{lipsum, color}

\let\vec\mathbf

\begin{document}
\title{Hierarchical Metric Learning for Optical Remote Sensing Scene Categorization}

\author{Akashdeep~Goel,
        Biplab Banerjee,~\IEEEmembership{Member,~IEEE}, and Aleksandra~Pi\v{z}urica,~\IEEEmembership{Senior Member,~IEEE}% <-this % stops a space
\thanks{A. Goel and B. Banerjee are with the Dept. of Computer Science and Engineering, Indian Institute of technology Roorkee, India. e-mail:\{akashdeep.goel1996, getbiplab\} @gmail.com. }% <-this % stops a space
\thanks{A. Pi\v{z}urica is with the Department of Telecommunications and Information
Processing,  Ghent University,  9000  Ghent, Belgium. e-mail: Aleksandra.Pizurica@UGent.be. }
}


\markboth{IEEE Geoscience and Remote Sensing Letters}%
{Shell \MakeLowercase{\textit{et al.}}: Bare Demo of IEEEtran.cls for Journals}



\maketitle

\begin{abstract}
We address the problem of scene classification from optical remote sensing (RS) images based on the paradigm of hierarchical metric learning. Ideally, supervised metric learning strategies learn a projection from a set of training data points so as to minimize intra-class variance while maximizing inter-class separability to the class label space. However, standard metric learning techniques do not incorporate the class interaction information in learning the transformation matrix, which is often considered to be a bottleneck while dealing with fine-grained visual categories. As a remedy, we propose to organize the classes in a hierarchical fashion by exploring their visual similarities and subsequently learn separate distance metric transformations for the classes present at the non-leaf nodes of the tree. We employ an iterative max-margin clustering strategy to obtain the hierarchical organization of the classes. Experiment results obtained on the large-scale NWPU-RESISC45 and the popular UC-Merced datasets demonstrate the efficacy of the proposed hierarchical metric learning based RS scene recognition strategy in comparison to the standard approaches.
 \end{abstract}

\begin{IEEEkeywords}
Optical remote sensing, metric learning, max-margin clustering.
\end{IEEEkeywords}

% \fi
\IEEEpeerreviewmaketitle

\section{Introduction}

\IEEEPARstart{R}{ecent} years have witnessed the continuous generation of satellite onboard remote sensing images which are characterized by fine spectral resolution and short revisit time \cite{brandtberg2006high}. Such images are used to capture the dynamics of the Earth's surface and hence aid in various applications including disaster management, urban planning and mineral studies, to name a few. 

Scene classification from very high resolution (VHR) optical RS imagery refers to the task of assigning unique semantic labels (e.g. parking lot, residential areas) to the scenes as a whole. Given the high spatial resolution, individual pixels of a VHR RS scene carry little information, in contrast to RS images with medium to low resolution (spatial resolution $\geq 30$m) where a given pixel represents a substantial area on ground. Hence, it is important to analyze the VHR RS images at the region or scene level and not only at the pixel level for the purpose of information extraction.

However, scene classification from VHR RS images is a complex task at its core given the varied nature of the ground terrains, differences in sensor viewpoints during image acquisition and radiometric image degradation due to atmospheric effects. This, in turn, causes substantial variations in the extracted feature descriptors from the images leading to an overlapping feature space. The performance of the standard classifier system is severely affected in such a scenario since it is difficult to model the class separators in the overlapping regions of the feature space. One of the popular solutions in this regard relies on learning a discriminative distance metric space from the original feature space where the data samples from different classes can be separated to the extent possible irrespective of the overlapping nature of the original feature space.

In particular, the goal of metric learning is to adapt some pairwise Mahalanobis distance metric function $d_\mathbb{M}(\mathbf{x},\mathbf{x}')=\sqrt{(\mathbf{x}-\mathbf{x}')^T\mathbb{M}(\mathbf{x}-\mathbf{x}')}$ for a given pair of samples $\mathbf{x}$ and $\mathbf{x}'$ to the problem of interest (supervised classification in our case) by leveraging the available training samples. Here $\mathbb{M}$ defines the symmetric, positive semi-definite metric projection matrix which is to be discriminatively learned. Broadly, the metric learning algorithms can be supervised, weakly-supervised or semi-supervised in nature. While supervised metric learning strategies explicitly make use of the label information in learning $\mathbb{M}$, weakly-supervised techniques rely on the indirect \texttt{must-link} or \texttt{no-link} constraints for a given pair of data samples. Although the supervised metric learning techniques make use of the label information in order to ensure maximum separation among the classes in the metric space, they largely ignore the visual relatedness among the classes in in modeling $\mathbb{M}$. Strictly speaking, an $\mathbb{M}$ learned from the samples of visually highly distinct classes usually fails to generalize well to a set of fine-grained categories. The problem is particularly of interest in case of scene recognition from optical RS data given that many of the scene themes are semantically related in general, e.g. sparse and dense residential.

In such a scenario, we advocate the need to learn separate $\mathbb{M}$s for different non-overlapping subsets of classes by exploring their semantic relatedness. This further leads to three distinct sub-problems from the point of view of the supervised classification task: i) organizing the classes into different subsets based on visual similarities, ii) performing separate metric learning for each subset, and iii) learning a sequence of classifiers that can exploit different metric spaces while performing inference. 

Based on these considerations, we propose a hierarchical supervised metric learning model in order to accomplish the task of RS scene recognition. The hierarchical model considered in this case is a binary tree structure which automatically divides the scene themes into different subsets from the root node (containing all the classes) to the leaf nodes (containing individual classes). In particular, the classes present at a given non-leaf node are divided into two finer sub-groups in order to construct the tree. Metric learning is subsequently adopted in each non-leaf node with the aim to maximize the separation between the two children of the node. In this way, the similarities among the classes are incorporated in learning a number of metric spaces at different levels of abstractions. A non-leaf node-specific binary classifier is further learned in the metric space for the purpose of separating its children. Classification for a test sample is performed by following the sequence of binary classifiers from the root to the leaf nodes of the tree. 

The main contribution of the proposed framework is the notion of hierarchical metric learning for fine grained RS scene recognition in a supervised context. To the best of our knowledge, this is one of the foremost endeavors which explores the notions of similarities among the classes and the idea of distance metric learning in designing an improved scene recognition system. Extensive experiments on the large-scale NWPU-RESISC45 dataset \cite{cheng2017remote} and UC-Merced \cite{yang2010bag} clearly show the efficacy of the proposed hierarchical metric learning based scene recognition strategy compared to standard baseline cases.

\section{Related works}

Considering the focus of the letter, we review briefly regarding: i) scene classification from optical RS images, and ii) metric learning techniques in the context of classifier learning.

\textbf{Classification of optical RS images}: With the availability of abundance of large-scale VHR RS image databases including UC-Merced \cite{yang2010bag}, AID \cite{xia2017aid}, and NWPU-RESISC45, the task of scene recognition has gained enormous popularity in the recent past \cite{cheng2017remote}. Considering the fact that the performance of a given visual recognition system heavily depends upon the discriminativeness of the underlying feature representations, the low, mid, and high level feature descriptors have been used for the scene recognition task to date. Amongst the ad-hoc low-level local features, SIFT, SURF, HOG are used for the same based on the paradigm of keypoints matching. However, each of these low-level descriptors alone lacks sufficient  generalization capabilities and the ability to adapt to major image transformations. As a remedy, the low-level descriptors are combined based on feature encoding strategies including bag of visual words, super-vector encoding (VLAD and Fisher's vector), sparse encoding (LLC), to name a few \cite{chatfield2011devil}. Thanks to the overwhelming success of deep learning techniques in visual inference tasks, deep Convolutional Networks (CNN) based feature descriptors are used in conjunction with RS data. Pre-trained CNN models including AlexNet, GoogleNet, VGGNet \cite{simonyan2014very}, and contextual CNN \cite{lee2017going} demonstrate excellent results in scene recognition. 

\textbf{Metric learning}: The goal of metric learning is to adapt a real-valued pairwise metric function to the classification problem in such a way that samples belonging to a given class come closer while samples from different classes are moved apart in the metric space. Ideally, the task is to learn a positive semi-definite transformation matrix from the feature space to the anticipated metric space such that the basic properties of pseudo-distance in the metric space: non-negativity, identity, symmetry, and triangular inequality, are preserved. A comparative analysis of different metric learning strategies is beyond the scope of this paper. Interested readers may consult \cite{kulis2013metric}.

Metric learning techniques have also been used in the analysis of remote sensing data. The motivation has mainly been into learning a discriminative feature space in order to deal with the mixed-pixel problem for RS image classification, e.g., the approach of \cite{pasolli2016active} combines large margin nearest neighbor (LMNN) \cite{weinberger2006distance} based dimensionality reduction and active learning based image classification for hyper-spectral data in a unified framework. In \cite{yang2017discriminative}, metric learning is employed for learning discriminative properties of hyper-spectral images in spatial and spectral domains. Considering the essence of contextual information for RS image classification, \cite{penga2014spatial} introduces a spectral-spatial metric learning strategy for hyper-spectral images considering the neighborhood information. Apart from image classification, metric learning has also been used for the purpose of target detection from hyper-spectral images \cite{dong2015maximum}.

The proposed framework shares some ideas with \cite{banerjee2017hierarchical} in the sense that we also follow the binary tree structure of the classes and a sequence of binary classifiers for inference. However, \cite{banerjee2017hierarchical} is focused to the problem of cross-domain classification of RS data following a domain generic subspace learning whereas we are interested in single domain classification. Moreover, we explore the notion of hierarchical metric learning in case of supervised classification.

\section{Proposed Framework}

\textbf{Problem definition}: Given $N$ image-label pairs $\chi_{TR}=\{(\mathbf{x_i},y_i)\}_{i=1}^N$ with $\mathbf{x_i} \in \mathbb{R}^d$ from $M$ categories ($y_i \in \{1,2,\ldots, M \}$), we aim at solving the following sub-tasks towards accomplishing the goal of scene recognition:
\begin{itemize}

\item Organize the $M$ classes in  a binary tree structure by exploring their visual features where each non-leaf node contains a subset of classes whereas the leaf nodes denote the individual ones.
\item Perform metric learning for each non-leaf node such that the separation between the two children of the node is maximized. 
\item Learn a (non)-leaf node specific binary classifier to distinguish between its children in the metric space (Figure 1).
\end{itemize}

\begin{figure*}
\centering
\includegraphics[scale=0.35]{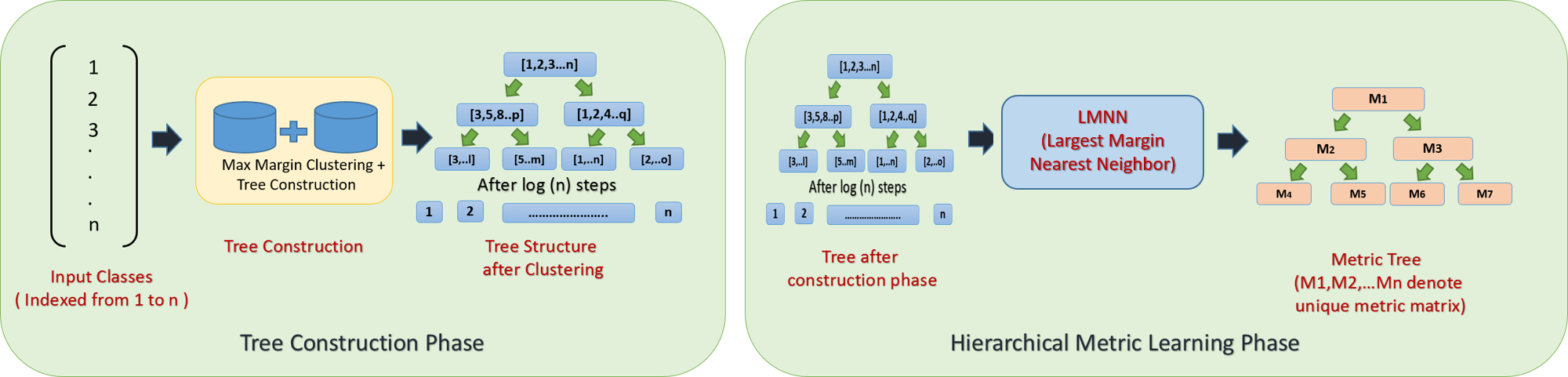}
\caption{A flowchart depicting the training stage of the proposed approach - class hierarchy construction and node specific metric learning}
\end{figure*}

\subsection{Building a hierarchical binary tree structure of the visual categories using maximum-margin clustering }
The goal of this stage is to organize the RS scene classes in a hierarchical binary tree fashion by exploring their visual features from $\chi_{TR}$. Given the representative samples for the classes, the clustering stage iteratively divides them into two finer groups thus building a binary tree structure. Notice that we consider the visual centroids of the classes for this clustering stage. For the sake of convenience, we denote the centroids as $\{\mathbf{x_j^c}\}_{j=1}^M$ from now onwards.
Since the visual features are overlapping in nature for a number of land-cover categories, the use of centroids as the representative is well-justified over the case of clustering all the samples together (the standard clustering process) which is largely affected by mis-classification and thus does not serve our purpose.

The literature for solving the clustering problem is rich \cite{aggarwal2013data}. However, given the small scale size of our dataset which is ideally the number of classes, we deploy notion of the maximum-margin clustering (MMC) \cite{zhang2009maximum}.
In particular, we consider the iterative support vector regression (SVR) based formulation \cite{zhang2009maximum} for the same where a constraint on the class balance is imposed while separating the data using a large-margin hyperplane to avoid any trivial solution. Henceforth, the standard binary clustering problem using MMC at the root node of the binary tree (considering all the classes) is formulated as follows:

Given $\chi_{TR}$ with $y_j \in \{+1,-1\}$ (since all the $M$ classes are to be divided into two sub-groups at the first level of the tree), the standard SVM classifier $(\vec{w},b)$ seeks to obtain the maximum-margin  hyperplane $f(\mathbf{x^c})=\mathbf{w}^T \phi(\mathbf{x^c})+b$ in some non-linear feature space $\phi$ by solving the following convex quadratic optimization problem in the primal:

\begin{equation}
\centering
\underset{\vec{w},b,\xi}{\mathrm{min}} ||\vec{w}||^2+2 C \xi \vec{e}
\end{equation} 

s/t,

\begin{equation}
\centering
y_j (\vec{w} \phi(\mathbf{x_j^c})+b) \geq 1- \xi_j
\end{equation}

for non-negative slack variables $\xi_j \geq 0$, regularization parameter $C >0$ and a vector $\vec{e}$ consisting of \texttt{ones}. 
Since the $y_j$s are ideally unknown initially in the unsupervised framework, a trivial solution assigns same class labels to all the samples resulting in an infinite margin.
As a remedy, a class-imbalance constraint is considered which emphasizes the $y_j$s to satisfy the following constraint for a non-negative trade-off parameter ($l \geq 0$):

\begin{equation}
\centering
-l \leq \vec{e}^T y \leq l
\end{equation}Following \cite{zhang2009maximum}, we solve this problem by using Laplacian loss based support vector regression (SVR) model. 
As already described, once we obtain two different sub-group of classes by applying MMC on all the classes at the root node, the same process is repeated separately for each of the children until the leaf nodes are reached.

Ideally, we follow the standard linear list based tree data structure in storing the binary tree where the children of the $i^{th}$ nodes are placed in the locations $2*i$ and $(2*i+1)$, respectively. Since the structure of the tree is largely dependent on the pairwise similarities among the classes, it is possible to obtain a tree which is skewed.

%\begin{algorithm}[!h]
%    \begin{algorithmic}[1]
%       \Function {Build}{$node,features,classes,size$} 
%        \State tree[node] = classes
%        \If {$size != 1$} \Comment{We have reached a leaf}
%        	\State [left,right] = MaxMarginCluster(features,classes)
%            \State leftFeatures = GetFeatures(left)
%            \State rightFeatures = GetFeatures(right)
%            \State Build(2*node+1, leftFeatures, left, size(left))
%            \State Build(2*node+2, rightFeatures, right, size(right))
            
%        \EndIf
%        \EndFunction
            
%    \end{algorithmic}
%    \label{alg:BuildTree}
%    \caption{Tree Construction}
% \end{algorithm}

\begin{algorithm}[!h]
    \begin{algorithmic}[1]
        \Function {DoLMNNTrain}{$node,features,tree$} 
        \State currentNode = tree[node]
        \State count = 0
        \If {size(currentNode) != 1}
        	\State leftChild = tree[2*node+1]
            \State rightChild = tree[2*node+2]            
        	\For {i = 1 to size(leftChild)} \Comment{No. of Classes}
        		\For {j = 1 to size(leftChild[i])} \Comment{Samples}
                	\State trainingLabel[count] = 1
                    \State count = count + 1
                \EndFor
        	\EndFor
            \For {i = 1 to size(rightChild)} \Comment{No. of Classes}
        		\For {j = 1 to size(rightChild[i])} \Comment{Samples}
                	\State trainingLabel[count] = 2
                    \State count = count + 1
                \EndFor
        	\EndFor
            \State [metric,details] = LMNN(features,trainingLabel)
            \State metricTree[node] = metric
            \State DoLMNNTrain(2*node+1,features,tree)
            \State DoLMNNTrain(2*node+2,features,tree)
		\EndIf
        \EndFunction
            
    \end{algorithmic}
    \label{alg:BuildLMNNTree}
    \caption{Hierarchical Metric Learning}
\end{algorithm}

\subsection{Hierarchical Metric Learning using LMNN }
%\begin{figure*}[t]
%\centering
%  \includegraphics[scale=0.45]{classification.png}
%  \caption{Image Classification Phase: Extracting deep CNN %features from image followed by metric space conversion. %These latent space features are then used for tracing a %path from root to a unique leaf node.}
%  \label{fig:pipeline}
%\end{figure*}

Once the hierarchical representation of the classes is obtained from a coarse to fine scale, a (non-leaf) node specific metric learning is carried out for better separation of the children of a given (non-leaf) node in the induced space, which may not be possible in the original feature space. We rely on LMNN based pseudo metric learning technique in this regard given its simplicity and prior successful applications in the area of RS \cite{penga2014spatial}.

The main idea behind LMNN is to learn a Mahalanobis metric under which all data instances in the training set are surrounded by at least $k$ non-overlapping samples sharing identical class labels. For a given sample, the target samples (with same label) should be close while drifting apart the impostors (samples with different labels). The final optimization problem considered for LMNN can be formulated as:
\begin{equation}
{\displaystyle \min _{\mathbf {\mathbb{M}} }\sum _{i,j\in \mathbf{N}_i} (\vec{x}_i-\vec{x}_j)^T\mathbb{M}(\vec{x}_i-\vec{x}_j)+\sum _{i,j,l}\zeta _{ijl}} 
\end{equation}
$ {\displaystyle \forall l, y_{l}\neq y_{i}}$, the non-negative slack variables $\zeta$ and the positive semi-definite projection matrix $\mathbb{M}$. $\mathbf{N}_i$ denotes the neighborhood for sample $i$. In addition, the following constraints are imposed to maintain the pre-defined fixed margin of $1$ unit between the classes:
\begin{gather}
    {\displaystyle d(\mathbf{x}_i, \mathbf{x}_j) + 1 \leq d(\mathbf{x}_i, \mathbf{x}_l) + \zeta_{ijl}} 
 \end{gather}We solve the problem using the traditional semi-definite programming strategy only based on the samples from the classes of the separate non-leaf nodes exclusively. As a result, we obtain a different metric projection matrices at the non-leaf nodes of the tree. 
In contrast to having a single metric for all the classes, we can now focus on the subset of classes with high appearance similarity and learn a discriminative metric space for them (Algorithm 1).
 
\subsection{Testing} During generalization, a test sample is fed to the root node of the tree and it follows the sequence of binary classifiers (standard KNN in our case) in the node specific learned metric spaces  before being assigned a label in one of the leaf nodes of the tree. We rely on KNN in this respect mainly for two reasons:

\begin{table*}[t]
\centering
\caption{Summary of results on both the datasets}
\begin{tabular}{|c|c|c|c|c|c|}\hline
\diaghead{\theadfont Diag ColumnmnHead II}%
{Train Percentage}{Methods}&
\thead{kNN\\ Classification}&\thead{Single level KNN with \\Metric Learning}&\thead{Hierarchical w/o \\Metric Learning}&\thead{Hierarchical with unique \\ Metric Learning }&\thead{Hierarchical with \\Metric Learning}\\    
\hline
50\% (NWPU-RESISC45) & 67.18\% & 70\% & 66.66\% & 58.97\% & \textbf{82.4\%}\\
\hline
80\% (NWPU-RESISC45) & 70.95\% &72\% & 77.4\% & 59.33\% & \textbf{84.6\%}\\
\hline
50\% (UC-Merced)& 87\% &88\% & 88\% & 82\% & \textbf{91\%}\\
\hline
80\% (UC-Merced)& 91\% &92\% & 91\% & 85\% & \textbf{94\%}\\
\hline
\end{tabular}
\end{table*}

%\begin{figure}
%\centering
%\includegraphics[scale = 0.4]{fig.jpg}
%\caption{Row 1: Within class variance (Beach), Row 2: %Between class similarity (Freeway, Overpass, %Intersection).}
%\label{fig:results}
%\end{figure}

\begin{itemize}
\item Ideally, once a good metric is learned, the problem of classification can simply be posed as the nearest neighbor searching.
\item LMNN is implicitly designed to work with the nearest neighbor classifier. So the task of constructing a number of node specific binary classifiers which is costly, can be alleviated.
\end{itemize}

Notice that the time complexity during training of the proposed approach is proportional to $\mathcal{O}(\ln M + M^2 + \ln M N^3) \approx \mathcal{O}(M^2 + N^3)$ for $N$ training samples and $M$ classes considering tree construction ($\mathcal{O}(\ln M)$ being the maximum depth of the tree), MMC ($\mathcal{O}(M^2)$), and node wise metric learning ($\mathcal{O}(N^3)$). Although semi-definite programming is rather time-consuming, the LMNN algorithm can be solved quite efficiently since most of the imposter constraints can be overlooked in general as they are obvious.
While during testing, the time required is proportional to $\mathcal{O}(\tau \ln M)$ where $\tau$ denotes a constant depicting the processing time per node.

\section{Results}

\subsection{Dataset Used and Experimental Setup}
We evaluate the efficacy of the proposed framework on two datasets: NWPU RESISC45 and UC-Merced, both of which are described in detail in an online repository \footnote{https://sites.google.com/view/zhouwx/dataset}. NWPU RESISC45 contains $31,500$ images depicting VHR scenes of man-made objects and typical land-cover themes from $45$ different classes and each class contains $700$ samples in total. On the other hand, UC-Merced consists of $2100$ images from $21$ land-cover classes ($100$ image per class). For experimental purpose, we consider two training-test data splits: 80\%-20\% and 50\%-50\%, respectively where we randomly and separately sample each class to construct the training and test sets. Note that we represent the images in terms of the $4096$ dimensional VGG-16 features extracted from $fc-6$.

The same experimental setup is followed for both the datasets. For LMNN metric learning, we consider $K=\{1,3,5,7\}$ during training the model and fix $K=7$ based on cross-validation.
Likewise during testing, we report the KNN classification performance for $K=\{1,3,5,7\}$. Surprisingly, we find that the classification performance during testing remains unchanged for different values of $K$ for both the datasets. This can be attributed to the discriminative feature spaces learned as a result of the per node metric learning strategy.
For the sake of comparison, we consider three benchmark scenarios:
1) standard single level multi-class KNN classifier 2) single level LMNN based KNN classifier 3) KNN with the proposed binary tree based hierarchy, and 4) KNN based hierarchical classification using a single metric learned from all the classes.

\subsection{Results and Discussion}

\begin{figure*}[t]
\centering
\includegraphics[scale=0.30]{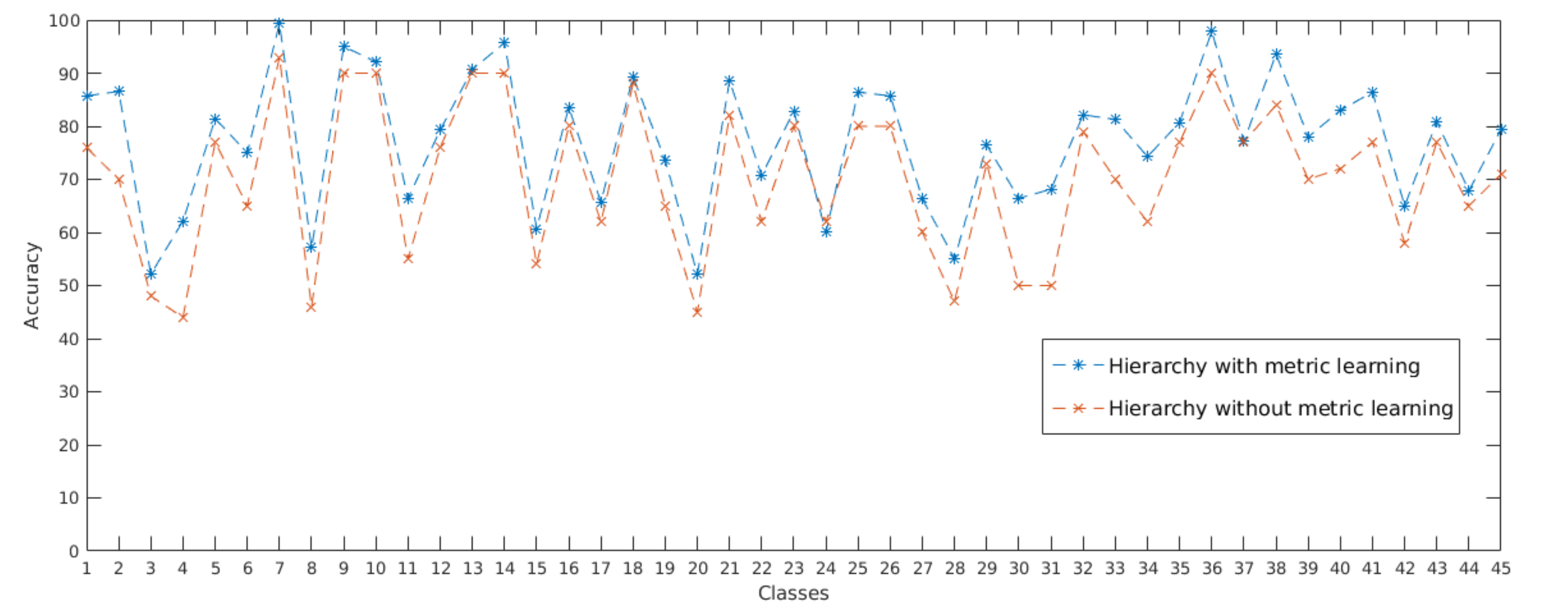}
\caption{Comparison of classwise performance (ordering of the classes (1-45) can be found in \cite{cheng2017remote}.}
\end{figure*}

Table 1 depicts the performance measures obtained from all the aforementioned classification frameworks. While experimenting on the $80-20$ train-test split, the standard KNN without metric learning outputs a mean classification accuracy of $\approx 71$ for NWPU-RESESC45. The performance is superior while the hierarchical binary tree based classification model is adopted. In particular, we obtain a classification performance of $\approx 77 \%$ while a hierarchical classification setup is considered without the application of LMNN based metric learning. On the contrary, we perform both single level and hierarchical classification based on the metric learned considering all the categories together. While the standard LMNN based single level classification yields a recognition performance of $\approx 72 \%$, the use of a single metric learned once considering all the classes at each non-leaf node of the tree produces a classification performance of $59 \%$. On the other hand, we observe a sharp rise in the classification accuracy when the proposed hierarchical metric learning based classification strategy is adopted. In particular, we obtain a mean classification performance of $\approx 85 \%$ for NWPU-RESESC45 dataset. 

Out of the $45$ categories present in the dataset, many of the classes share similar local geometrical structure: dense residential and commercial area, meadow and forest, to name a few. The classification performance on those classes are substantially low while the standard classification strategies are adopted without metric learning. Further, the use of a global metric considering all the classes fails to capture the class distributions at the finer level. Significant improvements in the recognition performance for such classes are observed ($ \geq 3-5 \%$) with the proposed hierarchical metric learning setup. Figure 2 depicts the classwise accuracy measures of the hierarchical classification framework (both with and without metric learning). It is evident from the measures that the proposed approach enhances the classification performance almost all the classes.

Similar trend is observed for UC-Merced as well where an overall classification performance of $94 \%$ is reached by the proposed framework which is better than all the cases considered on the $80-20$ training test split. Overall, all the techniques used for comparison produce high accuracy measures for this dataset. 

We also consider SVM coupled with linear kernel function and random forest classifier with $100$ component trees for per-node binary classification. While we observe comparable performance on both the datasets while SVM is used ($84 \%$ and $96 \%$ for NWPU-RESISC45 and UC-Merced, respectively on 80-20 split), the performance of random forest is worse by a margin of at least $10 \%$ which is due to model overfitting.

\section{Conclusion}
We propose a hierarchical metric learning based classification strategy for VHR optical RS scenes in this paper. In contrast to the standard metric learning approaches which are applied on the entire training set at once, we further explore the appearance relatedness of the scene categories in a hierarchical fashion and learn separate metric spaces on the subsets of visually similar classes. This helps in better classifying find-grained scene categories, which is reflected in the experiments. We are currently engaged in extending this framework for the purpose of cross-sensor remote sensing image classification.

% or


\section{Acknowledgement}
B. Banerjee was supported by SERB, Dept. of Science \& Technology, Govt. of India (ECR/2017/000365).


\ifCLASSOPTIONcaptionsoff
  \newpage
\fi



%\bibliographystyle{IEEEtran}
\bibliographystyle{IEEEtran}
\bibliography{bare_jrnl}



%\vfill


\end{document}